\RequirePackage{amsmath}
\documentclass[runningheads]{llncs}

\usepackage{graphicx}
\usepackage{amssymb}
\usepackage{amsmath}
\usepackage{upgreek}
\usepackage{dirtytalk}
\usepackage{url}
\usepackage{booktabs}
\usepackage{float}
\usepackage{bm}
\usepackage{tabularx}

\begin{document}

\title{
Pseudo-Random Number Generation
using Generative Adversarial
Networks}

\author{Marcello De Bernardi \inst{1} \and
MHR Khouzani \inst{1} \and
Pasquale Malacaria \inst{1}}

\authorrunning{M. De Bernardi et al.}

\institute{Queen Mary University of London, E1 4NS, UK \\
\email{m.e.debernardi@se15.qmul.ac.uk,
\{arman.khouzani,p.malacaria\}@qmul.ac.uk}}

\maketitle

\begin{abstract}
Pseudo-random number generators (PRNG) are a fundamental element of many security algorithms. We introduce a novel approach to their implementation, by proposing the use of generative adversarial networks (GAN) to train a neural network to behave as a PRNG. Furthermore, we showcase a number of interesting modifications to the standard GAN architecture. The most significant is partially concealing the output of the GAN's generator, and training the adversary to discover a mapping from the overt part to the concealed part. The generator therefore learns to produce values the adversary cannot predict, rather than to approximate an explicit reference distribution. We demonstrate that a GAN can effectively train even a small feed-forward fully connected neural network to produce pseudo-random number sequences with good statistical properties. At best, subjected to the NIST test suite, the trained generator passed around 99\% of test instances and 98\% of overall tests, outperforming a number of standard non-cryptographic PRNGs.

\keywords{adversarial neural networks  \and pseudo-random number generators \and neural cryptography}
\end{abstract}

\section{Introduction}

A \emph{pseudo-random number generator} (PRNG) is a deterministic algorithm with a secret internal state $S_i$ \cite[p. 2]{kelsey1998cryptanalytic}, which processes a random input seed $s$ to produce a large number sequence that may not tractably be distinguished by statistical means from a truly random sequence \cite[p. 170]{menezes1996handbook}. PRNGs are a fundamental element of many security applications \cite[p. 1]{kelsey1998cryptanalytic} \cite[p. 169]{menezes1996handbook}, where they are often a single point of failure, making their implementation a critical aspect of the overall design \cite[p. 2]{kelsey1998cryptanalytic}.


\subsubsection{Aims and Motivations}
The aim of this research is to determine whether a machine learning structure can learn to output sequences of numbers which appear randomly generated, and whether such a structure could be used as a PRNG in a security context. We confine this investigation to the statistical characteristics of a PRNG's output; cryptanalysis of the implementation, also necessary in order for a PRNG to be considered secure \cite{kelsey1998cryptanalytic,rukhin2001statistical}, is beyond the scope of this work. A statistically ideal PRNG is one that passes the \emph{theoretical next bit test} \cite[p. 171]{menezes1996handbook}.

The research is inspired by Abadi and Andersen's work on neural network learning of encryption schemes \cite{abadi2016learning}, conjecturing that a neural network can represent a good pseudo-random generator function, and that discovering such a function by stochastic gradient descent is tractable. Motivation is also drawn from the needs of security: a hypothetical neural-network-based PRNG has several potentially desirable properties. This includes the ability to perform ad-hoc modifications to the generator by means of further training, which could constitute the basis of strategies for dealing with the kind of non-statistical attacks described by Kelsey et al. in \cite{kelsey1998cryptanalytic}.

\subsubsection{Related Work}
Few attempts have been made to produce pseudo-random number sequences with neural networks \cite{desai2011pseudo,desai2012pseudo,tirdad2010hopfield,jeong2018pseudo}. The most successful approaches have been presented by Tirdad and Sadeghian \cite{tirdad2010hopfield}, and by Jeong et al. \cite{jeong2018pseudo}. The former employed Hopfield neural networks adapted so as to prevent convergence and encourage chaotic behavior, while the latter used an LSTM trained on a sample of random data to obtain indices into the digits of pi. Both papers reported a strong performance in statistical randomness tests. However, neither scheme sought to train an ``end-to-end" neural network PRNG, instead using the networks as components of more complex algorithms.

We undertake the task differently, by applying a deep learning method known as \emph{generative adversarial networks} \cite{goodfellow2014generative} to train an end-to-end neural PRNG which outputs pseudo-random sequences directly. We present two conceptually simple architectures, and evaluate their strength as PRNGs using the NIST test suite \cite{rukhin2001statistical}.

\subsubsection{Contributions}
This work makes a number of novel contributions to the field by proposing several modifications to the GAN framework. In summary, we introduce a simplification to the GAN framework that is applicable to this task, whereby the GAN does not include a reference dataset which the generator should learn to imitate. Furthermore, we also model the statefulness of a PRNG using a feed-forward neural network with supplementary non-random ``counter" inputs, rather than a recurrent network.

The overall product of these modifications is a system that is simple, conceptually elegant, and robust. We find that the trained generator can repeatedly pass approximately 98\% of NIST tests on default settings, showing that the adversarial approach is highly successful at training a neural network to behave as a PRNG. Our results are approximately on par with those of Tirdad and Sadeghian \cite{tirdad2010hopfield} and Jeong et al. \cite{jeong2018pseudo}, and outperform a number of standard PRNGs \cite{jeong2018pseudo}. Especially for a preliminary implementation, this outcome makes a strong case for further investigation.

\section{Design and Implementation}\label{chapter:design}

Let $\mathbb{B}$ be the set of all unsigned integers representable with 16 bits. For convenience we constrain the inputs and outputs of our networks to this range. We then view a pseudo-random number generator as any system implementing a function
\begin{equation}\label{eq:conceptual_prng}
prng(s) : \mathbb{B} \rightarrow {\mathbb{B}}^n
\end{equation}
where $s$ is a random seed, $n$ is very large, and the outputs of $prng$ fulfill some criteria for randomness. For individual outputs, we  can also characterize a PRNG as a function
\begin{equation}
prng^{\triangledown}(s, S_i) : X \rightarrow \mathbb{B}
\end{equation}
where $S_i$ is the current internal state of the generator, and $X$ is the set of all tuples $(s, S_i)$. 

\begin{figure}
\centering
\includegraphics[width=1\textwidth]{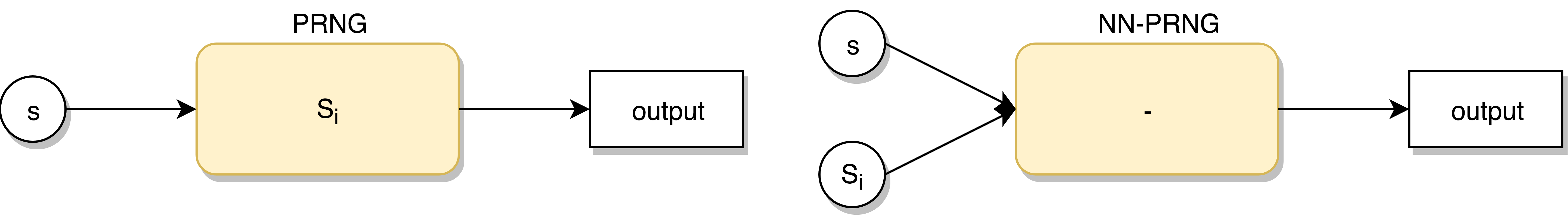}\\
\caption[PRNG black box vs. neural network PRNG black box]{Conceptual view of a PRNG (left) and our neural implementation (right).}
\label{figure:conceptual_difference}
\end{figure}

A generator neural network should represent a function $G(s)$ which approximates $prng(s)$. To simplify the design and training, we use a feed-forward (stateless) neural network, and model the PRNG's internal state $S_i$ as an additional $t$-dimensional input $\bm{o}_t$ instead (figure \ref{figure:conceptual_difference}). Thus the neural network actually represents a function
\begin{equation}
G^{\triangledown}(s, \bm{o}_t) : \mathbb{B}^{t+1} \rightarrow \mathbb{B}^n
\end{equation}
which approximates $prng^{\triangledown}(s, S_i)$, where $n$ is the network's output dimensionality. We can view $\bm{o}_t$ as an ``offset" into the full output sequence for $s$: for any fixed specific $s$, the complete pseudo-random sequence $G(s)$ is given by concatenating the generator's output sequences $\forall o_t\ G^{\triangledown}(s, \bm{o}_t)$. It follows that we have
\begin{equation}
|G(s)| \in \Uptheta(n^t)
\end{equation}
for the length of the full sequence $G(s)$.

\subsubsection{Adversarial Framework}
\begin{figure}
\centering
\includegraphics[width=0.99\textwidth]{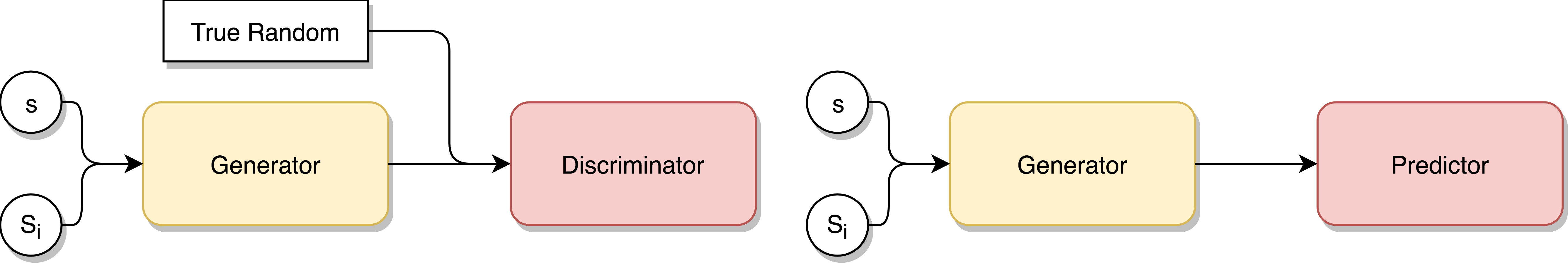}\\
\caption[Discriminative approach and predictive approach]{The discriminative approach (left) requires an external source of randomness which it attempts to imitate, while the predictive approach (right) has no external inputs.}
\label{figure:approach_comparison}
\end{figure}

A PRNG should minimize the probability of an adversary correctly guessing future outputs from past ones. This is analogous to a GAN, where the generator minimizes the probability of the discriminator accurately mapping its outputs to a class label \cite{goodfellow2014generative}. Thus we consider the generation of pseudo-random numbers as an adversarial task and formulate it using a GAN. We consider two distinct high-level architectures, termed the \textit{discriminative} and the \textit{predictive} architectures (figure \ref{figure:approach_comparison}). 

In the standard discriminative approach, the discriminator's inputs are number sequences drawn either from the generator or from a common source of randomness, and labeled accordingly. In order to minimize the probability of correct classification, the generator learns to mimic the distribution of the random sequences. 

In the predictive approach, loosely based on the theoretical next bit test, each sequence of length $n$ produced by the generator is split; the first $n - 1$ values are the input to the predictor, and the $n$th value is the corresponding label. The predictor maximizes the probability of correctly predicting the $n$th value from the other values, while the generator minimizes it. Thus the pseudo-randomness of the generator's output is formulated as unpredictability by an improving opponent.

\subsubsection{Generative Model}
\begin{figure}
\centering
\includegraphics[width=0.80\textwidth]{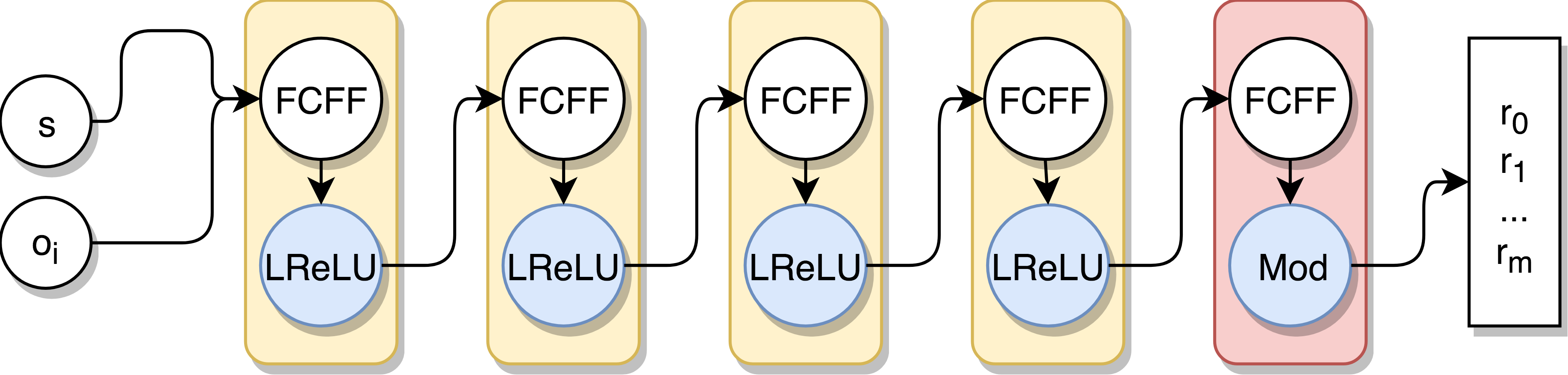}\\
\caption[Generator architecture]{Architecture of the generator: FCFF layers with leaky ReLU and mod activations}
\label{figure:architecture_generator}
\end{figure}

The generator is a fully connected feed-forward (FCFF) neural network representing the function 
\begin{equation}
G^{\triangledown}(s, o_1) : \mathbb{B}^2 \rightarrow \mathbb{B}^8.
\end{equation}
Its input is a vector consisting of a seed $s$ and a non-random scalar $o_1$ representing the PRNG state. It is implemented as four hidden FCFF layers of 30 units, and an output FCFF layer of 8 units (figure \ref{figure:architecture_generator}). The input layer and the hidden layers use the leaky ReLU activation function. The output layer applies $mod$ as an activation function, mapping values into a desired range while avoiding some of the pitfalls of $sigmoid$ and $tanh$ \cite[Neural Networks Part 1: Setting up the Architecture]{karpathy2017cs231n}.


\subsubsection{Discriminative Model}
\begin{figure}
\centering
\includegraphics[width=1\textwidth]{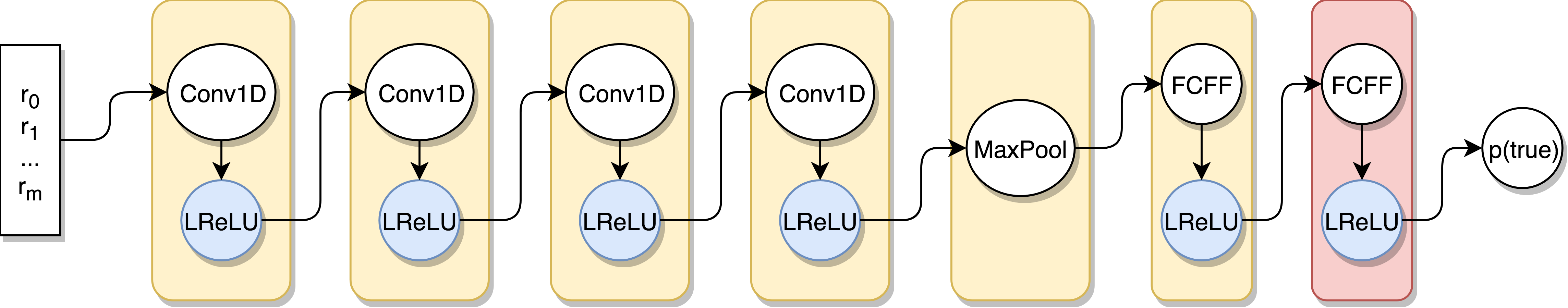}\\
\caption[Discriminator architecture]{Convolutional discriminator architecture. The output of the generator is convolved multiple times in order to extract higher-level features from the sequence; this is followed by pooling to reduce the output size, and FCFF layers to produce the final classification output.}
\label{figure:architecture_discriminator}
\end{figure}

The discriminator (figure \ref{figure:architecture_discriminator}) is convolutional neural network implementing the function
\begin{equation}
D(\bm{r}) : \mathbb{B}^8 \rightarrow [0, 1]
\end{equation}
where $\bm{r}$ is a vector of length $8$, either produced by the generator or drawn from a standard source of pseudo-randomness and associated with corresponding class labels. The discriminator outputs a scalar $p(true)$ in the range $[0, 1]$ representing the probability that the sequence belongs to either class.

The discriminator consists of four stacked convolutional layers, each with 4 filters, kernel size 2, and stride 1, followed by a max pooling layer and two FCFF layers with 4 and 1 units, respectively. The stack of convolutional layers allow the network to discover complex patterns in the input.

\subsubsection{Predictive Model}
The predictor is a convolutional neural network implementing the function
\begin{gather}
P(\bm{r}_{split}) : \mathbb{B}^{7} \rightarrow \mathbb{B}
\end{gather}
where $\bm{r}_{split}$ is the generator's output vector with the last element removed. The last element is used as the corresponding label for the predictor's input. Apart from the input size and meaning of the output, the discriminator and the predictor share the same architecture.


\subsubsection{Loss Functions and Optimizer}
We use standard loss functions. In the discriminative case, the generator and discriminator both have \emph{least squares} loss. In the predictive case, the generator and the predictor both have \emph{absolute difference} loss. We use the popular Adam stochastic gradient descent optimizer \cite{kingma2014adam}.

\section{Experiments}\label{chapter:experiments}
We measure the extent to which training the GANs improves the randomness properties of the generators by analyzing large quantities of outputs, produced for a single seed, using the NIST statistical test suite both before and after training. \\

\begin{itemize}
	\itemsep0em
    \item \textbf{Independent variable}: whether the GAN has been trained or not.
	\item \textbf{Dependent variable}: the result of the NIST tests.
	\item \textbf{Controlled variables}: the random seed, the non-random generator inputs, the architecture of the networks, and all training parameters such as number of epochs, learning rate, and mini-batch size, are fixed throughout the experiment.
\end{itemize}


\subsubsection{Experimental Procedure}
We initialize the predefined evaluation dataset $Data$ first. It consists of input vectors $\bm{v}_i \in \mathbb{B}^2$ of the form $[s, {o_1}_i]$, such that the random seed $s$ in $\bm{v}_i$ is fixed to the same arbitrary value for all $i$ and all experiments. The offset ${o_1}_i$ in $\bm{v}_i$ starts at 0 for $\bm{v}_0$ and increments sequentially for the following vectors. For example, assuming arbitrarily that $s=10$, we would have
\begin{equation}
Data=\bigl[[10, 0], [10, 1], [10, 2], ...\bigr]
\end{equation}

We use the untrained generator to generate floating-point output vectors for all vectors in $Data$. These values are rounded to the nearest integer. If the outputs are uniformly distributed over a range $[a,b]$ where $a,b \in \mathbb{R}^+$, then they will also be uniformly distributed over the range $[a,b]$ where $a,b \in \mathbb{Z}^+$. The integers produced are stored in an ASCII text file in binary format.

We then train the networks, with the generator and the adversary performing gradient updates in turn as is standard with GANs. The trained generator is used to produce another text file of output integers. The NIST test suite is executed on the files, enabling the evaluation of the generator's performance before and after training. For both the discriminative and predictive approaches, we carry out the procedure 10 times.

\subsubsection{Training parameters} 
In each experiment we train the GAN for 200,000 epochs over mini-batches of 2,048 samples, with the generator performing one gradient update per mini-batch and the adversary performing three. We set the learning rate of the networks to 0.02. The generator outputs floating-point numbers constrained to the range $[0, 2^{16}-1]$, which are rounded to the nearest 16-bit integer for evaluation. The evaluation dataset consists of 400 mini-batches of 2,048 input vectors each, for a total of 819,200 input samples. The generator outputs 8 floating-point numbers for each input, each yielding 16 bits for the full output sequence. In total, each evaluation output thus consists of 104,857,600 bits, produced from a single random seed. Larger outputs were not produced due to disk quotas on the cluster used to run the models.

\subsubsection{NIST testing procedure}
The NIST test suite is applied with default settings. The test suite consists of 188 distinct tests, each repeated 10 times, with 1,000,000 input bits consumed for each repetition. Each repetition will be referred to as a \emph{test instance}. For every test, NIST reports the number of individual instances that passed, the p-value of all individual instances, as well as a p-value for the distribution of the instance p-values. A test instance fails if its p-value is below a critical value ($\alpha = 0.01$). An overall test fails if either the number of passed instances is below a threshold, or the p-value for the distribution of test instance p-values is below a critical value.

\subsubsection{Results}
\begin{figure}[H]
    \centering
    \textbf{Training Loss, Discriminative Experiment 9}
    \includegraphics[width=0.9\textwidth]{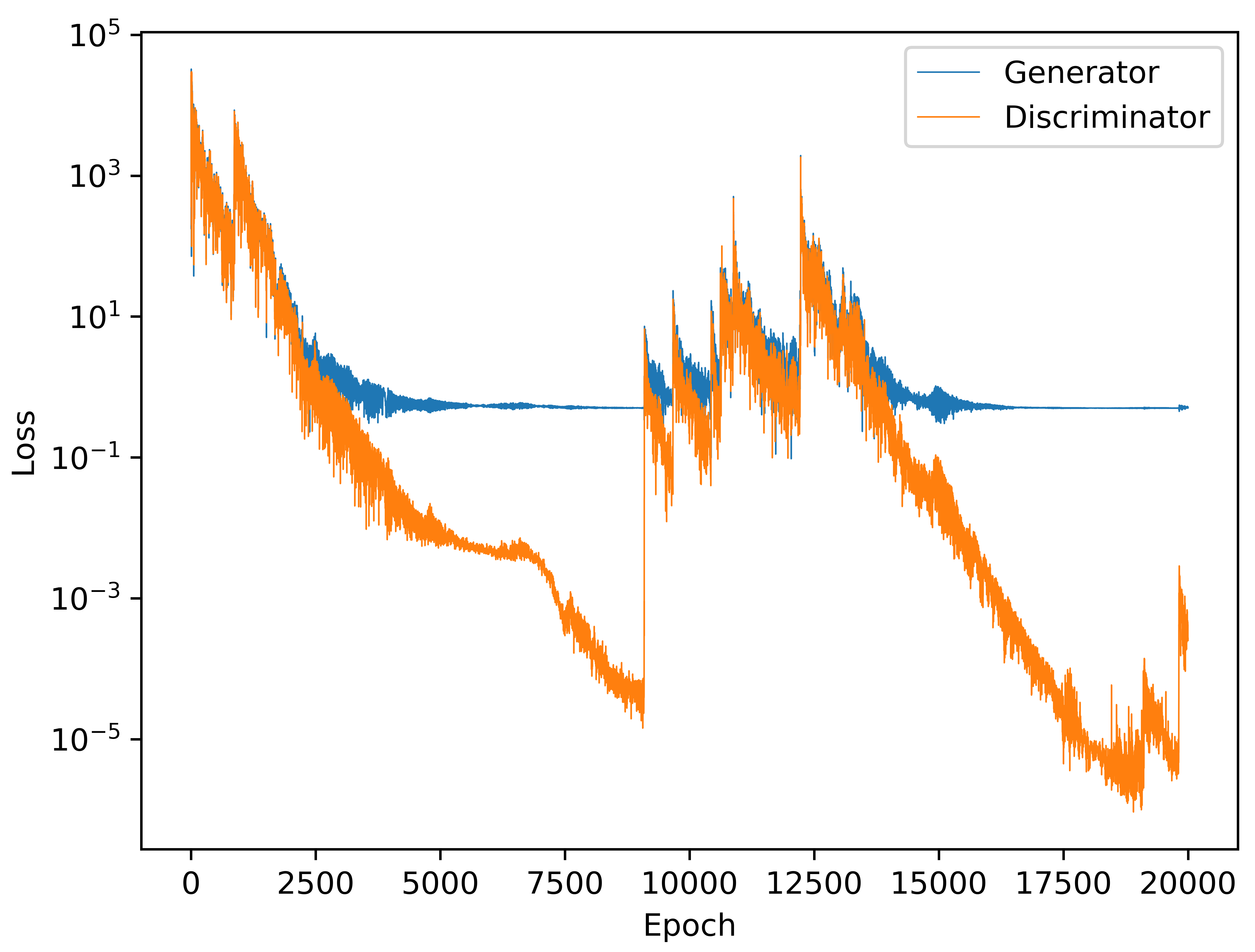}\\
    \caption[Training loss, discriminative experiment 9]{Training loss of the discriminative model. The discriminator has a tendency to gradually improve its performance while the generator plateaus. Occasionally the learning destabilizes and the discriminator's loss increases by a large factor.\\}
    \label{figure:loss_discgan}
    \centering
    \textbf{Training Loss, Predictive Experiment 9}
    \includegraphics[width=0.9\textwidth]{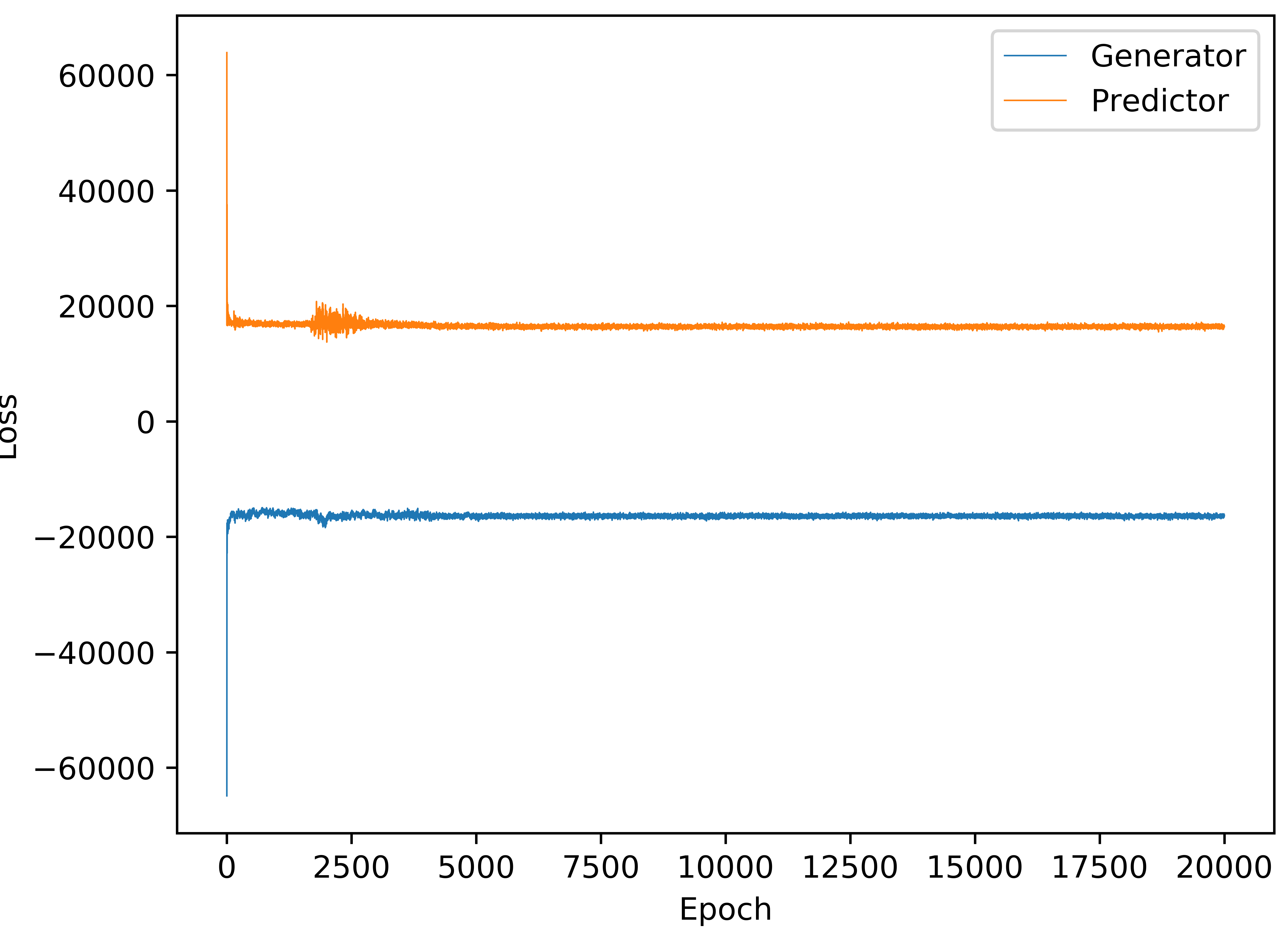}\\
    \caption[Training loss, predictive experiment 9]{A plot of the training loss during training of the predictive model. The predictor and generator converge in the initial phase of training.}
    \label{figure:loss_predgan}
\end{figure}

Table \ref{table:after_training_avg} shows the average performance across experiments, before and after training, for both GAN approaches. Table \ref{table:avg_result} shows the average improvement across all experiments for both approaches. Figures \ref{figure:loss_discgan} and \ref{figure:loss_predgan} display the loss functions during the a discriminative training run and a predictive training run.

\begin{table}
\caption[test results for trained generators]{ NIST test suite results for the generators, before and after training. $D_i$ and $P_i$ refer to discriminative and predictive experiments, respectively. $T$ is the overall number of distinct tests carried out by NIST STS, and $T_I$ is the number of total test instances. $F_I$ and $F_{I\%}$ are the number of failed test instances and the percentage of failed test instances. $F_p$ is the number of distinct tests failed due to an abnormal distribution of the test instance p-values. $F_T$ and $F_{\%}$ refer to the absolute number and percentage of distinct tests failed.}
    \begin{tabularx}{\textwidth}{lXXXXXXX} \toprule
    {$i$}   & {$T$} 	& {$\langle T_I \rangle$}	& {$\langle F_I \rangle$}   	& {$\langle F_{I\%} \rangle / \%$}		    & {$\langle F_p \rangle$} 	    & {$\langle F_T \rangle$} 	    & {$\langle F_{\%} \rangle / \%$} \\ \midrule
    $D_{before}$	& 188  & 1800		& 1796			& 99.8					& 188				& 188				& 100.0 \\ \midrule
    $D_{after}$     & 188  & 1800 		& 61	  		& 3.5					& 4.3		 		& 6.9		 		& 3.9 \\ \midrule
    $P_{before}$	& 188  & 1800		& 1798			& 99.9					& 188				& 188				& 100.0 \\ \midrule
    $P_{after}$     & 188  & 1830 		& 56  			& 3.0					& 2.7		 		& 4.5		 		& 2.5 \\ \bottomrule
\end{tabularx}
\label{table:after_training_avg}
\end{table}

\begin{table}
\caption[Average performance change across all experiments]{Performance change from before training to after training for the discriminative and predictive approaches across all tests.}
    \begin{tabularx}{\textwidth}{lXXXX} \toprule
    {$i$}     & {$\langle \Delta F_{I\%} \rangle / \%$}	& {$\langle \Delta F_p \rangle$} 	& {$\langle \Delta F_T \rangle$} 	& {$\langle \Delta F_{\%} \rangle / \%$} \\ \midrule
    $D$ & -96.2 						& -183.7					& 	-180.1	  					& -96.1 \\ \midrule
    $P$ & -96.7  						& -185.3					& -183.6						& -97.5 \\ \bottomrule
\end{tabularx}
\label{table:avg_result}
\end{table}

\subsubsection{Evaluation}
\begin{figure}[H]
    \centering
    \textbf{Output Sample, Before and After Predictive Training}
    \includegraphics[width=0.95\textwidth]{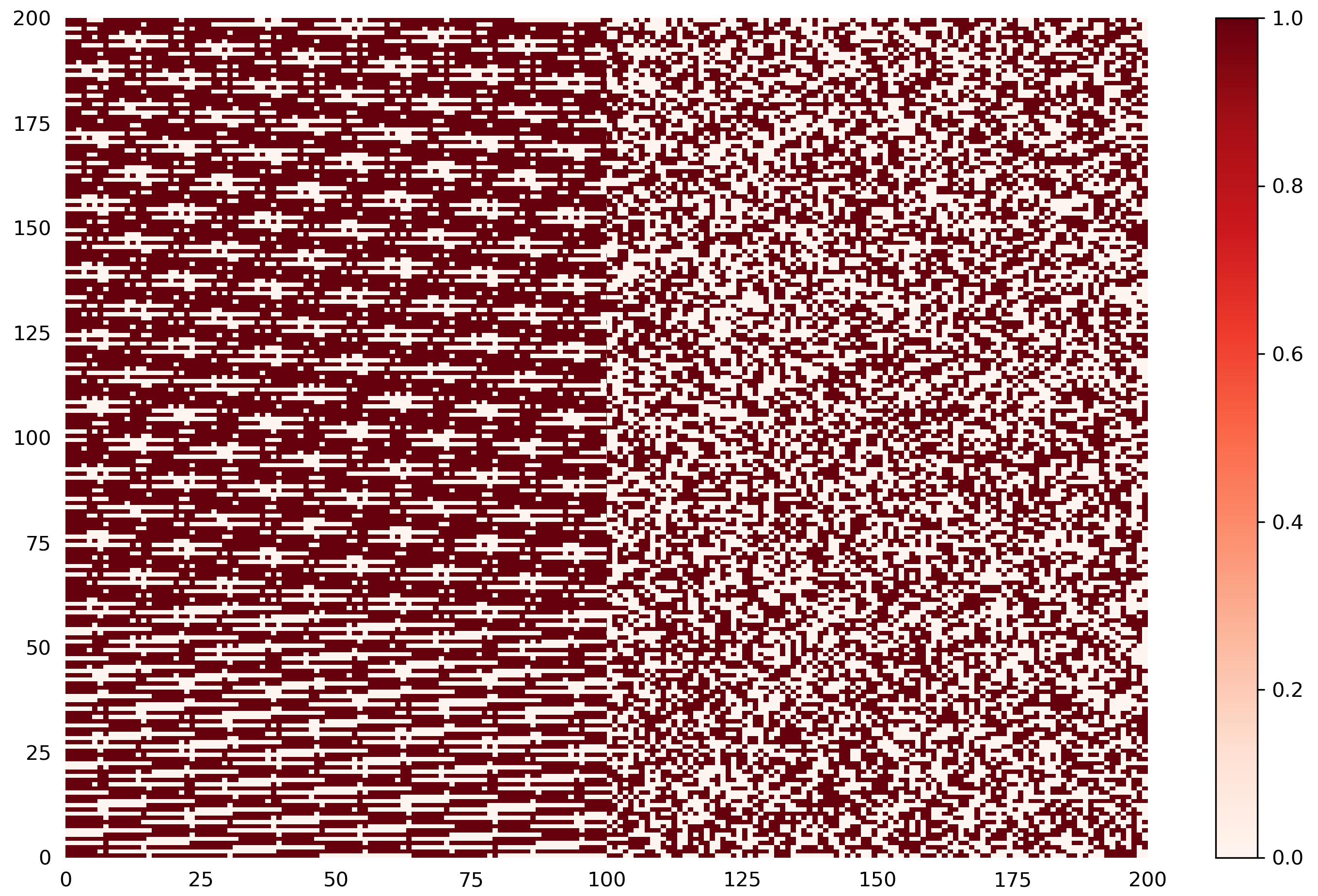}\\
    \caption[Visualization of output randomness before and after training, P9]{Visualization of the generator output as produced in the 9th predictive training instance, before (left half) and after (right half) training. The 200x200 grid shows the first 40,000 bits in the generator's sample output. Obvious patterns are visible before training, but not after.}
    \label{figure:visualize_predictive_before_after}
\end{figure}

Prior to training, the generators pass no statistical tests. After training the performance of the generators is consistently very strong for both approaches. The evaluated number sequences achieved a failure rate well below 5\% in the majority of experiments, with an absolute change in failure percentage as a result of training greater than 95\% in most cases. This on par with the results obtained by Tirdad and Sadeghian, whose best pass rate was around 98\% \cite{tirdad2010hopfield}. According to the data collected by Jeong et al., this also outperforms a number of standard non-cryptographic PRNGs. The difference in entropy of the output before training and after training is visualized in figure \ref{figure:visualize_predictive_before_after}.

The training loss plots are unusual. In the discriminative case (figure \ref{figure:loss_discgan}) we observe long periods of steady convergence, with short bursts of instability caused perhaps by the generator discovering a noticeably different pseudo-random function. The predictive case (figure \ref{figure:loss_predgan}) is characterized by very fast convergence during the first epochs, followed by long-term stability. An explanation could be a state of balanced learning, where both networks are improving together at a similar pace, maintaining their relative performance.

The predictive approach shows better results, with the generators producing approximately $60\%$ of the number of failures produced by the discriminatively trained generator. Moreover, we observed that training steps for the predictive GAN executed in about half the time.

\section{Conclusion and Further Investigation}\label{chapter:conclusion}

The aim of this investigation was to determine whether a deep neural network can be trained to generate pseudo-random sequences, motivated by the observation that GANs resemble the roles of a PRNG and an adversary in a security context. We explore a novel approach, presenting two GAN models designed for this task.

The design includes several innovative modifications applicable to this task. In particular, the predictive model waives the need for a reference distribution by making the desired distribution implicit in the adversarial game. Moreover, we forgo the use of recurrent architectures in favor of a feed-forward architecture with non-random ``counter" inputs.

We show that the adversarial approach is highly successful at training the generator. Training improved its performance significantly, resulting at best in passing around 99\% of test instances and 98\% of unique tests. To our knowledge, this is the first example of a neural net learning a PRNG function end-to-end.

We encourage further work to take a systematic approach to model selection and hyper-parameter optimization, and to investigate the learning process.

\bibliographystyle{splncs04}
\bibliography{references.bib}

\end{document}